\documentclass[twocolumn,twoside]{Jornadas}
\usepackage[utf8]{inputenc}
\usepackage{listings}
\usepackage{algorithm}
\usepackage{algorithmicx}
\usepackage{algcompatible}
\usepackage{adjustbox}
\usepackage{graphicx}
\usepackage{color}
\usepackage{caption}
\usepackage{amsmath}
\usepackage{amsfonts}
\usepackage{amssymb}
\usepackage[caption=false,font=footnotesize]{subfig}
\usepackage{placeins}
\usepackage{hyperref}
\usepackage{url}

\definecolor{gray97}{gray}{.97}
\definecolor{gray75}{gray}{.75}
\definecolor{gray45}{gray}{.45}

\def\BibTeX{{\rm B\kern-.05em{\sc i\kern-.025em b}\kern-.08em
    T\kern-.1667em\lower.7ex\hbox{E}\kern-.125emX}}

\graphicspath{{.}{./Figuras/}}

\begin{document}

\title{Operator Autoencoders: Learning Physical Operations on Encoded Molecular Graphs}

\author{%
     Willis Hoke%
     \thanks{Department of Computer Science, Portand State University},
     Daniel Shea%
     \thanks{Department of Materials Science and Engineering, University of Washington},
     and Stephen Casey%
     \thanks{NASA Langley Research Center; 
     e-mail: {\tt Stephen.Casey@nasa.gov}}
}

\maketitle
\markboth{}{}
\pagestyle{empty} 
\thispagestyle{empty} 

\begin{abstract}
Molecular dynamics simulations produce data with complex nonlinear dynamics. If the timestep behavior of such a dynamic system can be represented by a linear operator, future states can be inferred directly without expensive simulations.  The use of an autoencoder in combination with a physical timestep operator allows both the relevant structural characteristics of the molecular graphs and the underlying physics of the system to be isolated during the training process.   In this work, we develop a pipeline for establishing graph-structured representations of time-series volumetric data from molecular dynamics simulations.  We then train an autoencoder to find nonlinear mappings to a latent space where future timesteps can be predicted through application of a linear operator trained in tandem with the autoencoder.  Increasing the dimensionality of the autoencoder output is shown to improve the accuracy of the physical timestep operator.
\end{abstract}


\section{Introduction}
\PARstart{R}{easearch} into generating graph structures with desired properties has a history dating at least as far back as 1960 [1], with several diverse approaches having subsequently been developed related to statistical graph representation [2]-[5].    Prior work developing generative models for molecular structures [6] has demonstrated the usefulness of graph representations for volumetric data.  Concurrently, deep learning techniques have demonstrated notable success in various domains including imaging [7],[8], text [9],[10], and speech [11],[12].  Graph-related deep learning methods have been developed in recent years, focusing on areas such as graph generators [13], graph representation learning methods [14]-[18], graph attention models [19], and attack and defense techniques on graph data [20].

Graph autoencoders are neural networks trained to represent the structural features of a set of graph adjacency matrices as a vector of latent space variables [21]-[38].  Neural networks provide a powerful tool for nonlinear dimensionality transformation; in cases where input data do not lie on a linear manifold, an autoencoder can learn a mapping to a higher- or lower-dimensional space capturing the inherent structure of the data.  However, autoencoded adjacency matrices can become computationally intractable for graphs with large numbers of nodes [6].  To overcome this limitation for large molecular graphs we find the local graph representation in the neighborhood around each atom, thereby creating a set of overlapping subgraphs in 3D space.  Each subgraph is encoded into a corresponding latent vector.  

Graph decoding is the inverse of the encoding function whereby the latent vectors are reconstituted into adjacency matrices.  These adjacency matrices produce overlapping subgraphs which must undergo a graph-matching procedure in order to reconstruct the full volume.  Studies in graph matching have considered this problem from deep learning and combinatorial optimization perspectives [39]-[42].  To assist with reconstruction of our full-volume molecular graph, we introduce a canonical graph representation which eliminates the need for computationally expensive graph matching methods otherwise required to determine the similarity of overlapping graphs.

In addition to autoencoding, recent works have used deep neural networks to discover advantageous coordinate transformations for dynamic systems [43]-[50].  Such a transformation function can be modelled as linear Koopman operator acting on a state space of system observables [51]-[55].  Following results shown by [56], we train an autoencoder in parallel with a linear operator and demonstrate the use of high-dimensional latent representations to facilitate discovery of linear models for local system dynamics. Our model is trained on time-series data of diamond structures subjected to tensile deformation in the LAMMPS molecular dynamics simulation environment.

\section{Methods}

\subsection{Subvolume Sampling}
To learn approximations of global system dynamics, we consider local neighborhoods of atoms within a larger molecular structure. This has the dual advantage of shortening computation time caused by encoding large graphs while improving results in cases where local system dynamics are better approximated by a linear operator than global dynamics. By oversampling subsets of the initial dataset, we ensure graph representations of subvolumes can be projected back into three-dimensional coordinate space aligned to reconstruct the original volume.

Initial experiments utilized cubical subvolumes sampled along a grid. However, even in the optimal case this gives subvolumes of different sizes, which required zero-padding distance matrices after finding graph representations. Thus, a \textit{k}-nearest neighbors algorithm was applied to each point in the initial volume to generate $n$ subvolumes with $k$ points each. 

Since the LAMMPS [57] environment can use periodic boundary conditions for molecular dynamics simulations, we truncate the initial volumetric data to exclude atoms within a certain percent $p$ distance from the bounding box. We found $p = 5$ to be a sufficient value to eliminate boundary artifacts. This prevents sudden variations in atomic position between consecutive time steps.

\begin{figure}
\centering
  \includegraphics[scale=.8]{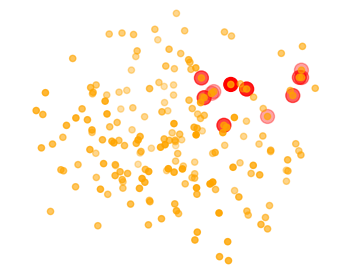}
  \caption{Example subvolume (red) found using $K$-Nearest Neighbors}
  \label{fig:subvol}
\end{figure}

\subsection{Distance Matrices \& Bond Order Potentials}
Graph representations have clear advantages over coordinate representations for learning latent embeddings. First, since they explicitly represent the relationships between atoms in a subvolume, they directly capture structural features. Secondly, since graph representations are invariant to linear or affine transformation, they provide a compact representation for volumes with similar structural features.

Pairwise distances are calculated between atoms in each subvolume using a standard Euclidean metric to yield the matrix $D \in {\mathbb{R}}^{k \times k}$. Reconstruction from distance matrices was demonstrated on randomized data using classical Multi-Dimensional Scaling (MDS) followed by Procrustes reconstruction.

Bond order potentials are next calculated from the pairwise distance matrix. We chose a two-body potential as three-body potentials require a significant number of parameters, some of which are unknown or unverified. The Lennard Jones potential is calculated as
\begin{equation}
V_{\text{LJ}}(r)=4\varepsilon \left[\left({\frac {\sigma }{r}}\right)^{12}-\left({\frac {\sigma }{r}}\right)^{6}\right]
\end{equation}

Here $\varepsilon$ and $\sigma$ are parameters chosen specifically according to atom types and molecular configuration. We used $\varepsilon = 0.7$ and $\sigma = 1.45$ based on experimental parameters for similar carbon structures [58].

\subsection{Canonical Graph Representations}
Past examples of graph autoencoders [6] have relied on approximation strategies like Max Pooling Matching (MPM)[59] to evaluate graph similarity. Since evaluation is computationally expensive and occurs in the loss function, training becomes intractable for larger molecule sizes. Our alternative approach involves preprocessing all graphs such that any permutation of the indices of the adjacency matrix maps to the same representation. We define the ordering mapping a graph to its canonical representation by ordering the indices of a distance matrix according to their magnitude:

\begin{equation}
    \begin{split}
    	\mathbf{p} = 
    	\begin{bmatrix}
    		i_1 & i_2 & \hdots & i_n
    	\end{bmatrix}
    	\: \mathrm{:} \:\:\:\:\:\:\:\:\:\:\:\:\:\:\:\:\:\:\: \\
    	\lVert{D_{(i_k)}}\rVert  <  
    	\lVert{D_{(i_{k+1})}}\rVert 
    	\ \forall \;
    	k \ \in \{1, 2, \dots, n - 1\}
	\end{split}
\end{equation}

\begin{figure}
  \includegraphics[width=\linewidth]{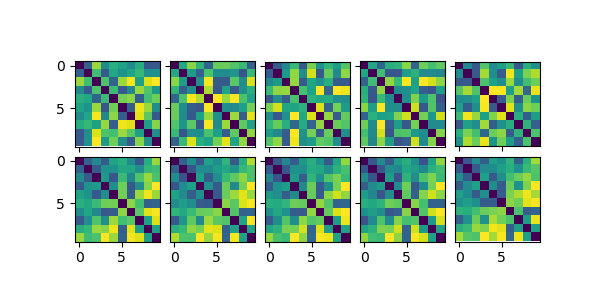}
  \caption{Similar distance matrices (top) with corresponding canonical representations (bottom)}
  \label{fig:canonical}
\end{figure}

The vector $\mathbf{p}$ uniquely determines the permutation applied to the rows and columns of a distance matrix. Not only does the canonical ordering map different representations of the same graph to the same representation, it also yields similar orderings for similar graphs. We tested this hypothesis by generating random permutations of the same distance matrix, adding a small amount of Gaussian noise, then computing canonical representations. As can be seen in Figure  
\ref{fig:canonical}, the canonical representation is invariant to permutation and resilient to small perturbations of pairwise distances, with the large majority of rows and columns remaining in the same canonical order.

\subsection{Dataset}
Our dataset was generated from a simulation of a diamond structure containing 1,000 carbon atoms subjected to tensile deformation over a total of 10 time steps sampled at intervals of 1,000. We selected $k = 10$ to generate subvolumes with 10 atoms each. Pairwise distances were computed using the standard Euclidean metric, then bond-order potentials were computed using the formula outlined above. Data were then seperated into pairs of vectors containing upper traingular entries from bond-order potential matrices at consecutive time steps. Finally, data was scaled to the range $[0,1]$ using a standard min-max scaling algorithm. The final training dataset contained 5,248 pairs of input vectors $v_{t_i}, v_{t_{i+1}} \in {\mathbb{R}}^{\frac{k(k-1)}{2}}$. A separate test dataset is not utilized since the goal is to learn a linear representation of system dynamics rather than generalizing to unseen data.

\subsection{Model Architecture}
The operator-autoencoder model consists of two networks trained in parallel. The first network is an multi-layer autoencoder, while the second represents a weight matrix defining a linear operator.

For the autoencoder, we chose a network with a total of four fully connected feed-forward layers, with the encoder and decoder each containing two layers. Hidden dimension $h$ for hidden layers in encoder and decoder was rounded to the nearest integer to the mean of the input and latent dimensions. Rectified Linear Units (ReLU) were used as activation functions between all layers except the final layer in the decoder, where we used a sigmoidal activation function to constrain output to the range $[0, 1]$.

The linear operator consists of a matrix $M \in {\mathbb{R}}^{d \times d}$, where $d$ is the dimension of the latent space. The initial value of the operator is set to the identity matrix $I_m$ as this preserves the initial representation of a subvolume and provides the closest known approximation to the next time step.

\subsection{Loss Functions}

We utilize two separate loss functions for model training. The first loss function quantifies reconstruction error from the autoencoder using Mean Squared Error (MSE) between an initial subvolume and its reconstruction. Reconstruction loss is defined as:
\begin{equation}
	{\mathcal{L}_{AE}} = \lVert \hat{u}^{(t_i)}_j - u^{(t_i)}_j  \rVert_2^2
\end{equation}

The second loss function provides a metric for accuracy of the time step operator. Operator loss is defined using MSE between the predicted and actual latent representation at time step $t + 1$:
\begin{equation}
	{\mathcal{L}_{OP}} = \alpha \lVert M(v^{(t_i)}_j)-v^{(t_{i+1})}_j \rVert_2^2
\end{equation}

Here $\alpha$ is a hyperparameter which determines the relative contribution of operator loss. Since operator loss backpropogates through the encoder, setting too small a value for $\alpha$ can result in the operator loss steadily increasing over time as the reconstruction loss dominates. 

\subsection{Training}
During each training epoch, mini-batches consisting of training data representing pairs $\left(u^{(t_i)}, u^{(t_{i+1})} \right)$ of subvolumes  at consecutive time steps pass through the autoencoder to yield latent representations $\left(v^{(t_i)}, v^{(t_{i+1})} \right)$ and reconstructions $\left(\hat{u}^{(t_i)}, \hat{u}^{(t_{i+1})} \right)$. The operator $M$ is applied to find $M(v^{(t_i)})$ during the forward pass. Reconstruction and operator loss are computed, then weights for the autoencoder network and operator matrix are updated through backpropogation. This affects the weights of both the operator and the encoder.

After hand-tuning hyperparameters to minimize loss values, we settled on a learning rate of $0.001$ with operator training delay of $50$ epochs. We fixed the operator loss coefficient $\alpha = 20$. The ADAM optimizer was used for training with a batch size of $64$. We trained a total of 9 models for 1500 epochs each. In contrast to typical autoencoder architectures which utilize a low-dimensional latent representation, we instead experiment with various dimension sizes ranging from small to large. For the purpose of training a linear operator, we find larger latent spaces can yield representations with lower operator loss.

\section{Results}

\begin{center}

 \begin{tabular}{| c | c c |} 
 \hline
 $d$ & $min\left(\mathcal{L}_{AE}\right)$ & $min\left(\mathcal{L}_{OP}\right)$ \\
 \hline
 16   & 0.2724 & 0.0013  \\
 32   & 0.1831 & 0.0026  \\
 64   & 0.1315 & 0.0058  \\
 128  & 0.0288 & 0.0040  \\
 256  & 0.0086 & 0.0024  \\
 512  & 0.0041 & 0.0012  \\
 1024 & 0.0030 & 0.0008  \\
 2048 & 0.0016 & 0.0005  \\
 4096 & 0.0005 & 0.0003  \\
 \hline
\end{tabular}

\bigskip

\textbf{Minimum loss values as a function of latent dimension}\\

\end{center}

\begin{figure}
  \includegraphics[width=\linewidth]{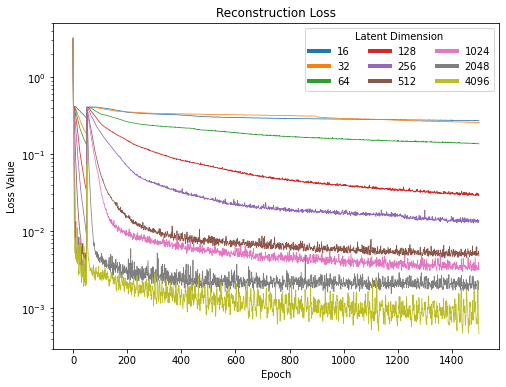}
  \caption{Reconstruction loss as a function of latent dimension}
  \label{fig:reconstruction_loss}
\end{figure}

Figure \ref{fig:reconstruction_loss} shows results for $\alpha$-scaled reconstruction loss for each model. In every case the introduction of the operator loss $\mathcal{L}_{OP}$ after a delay of $50$ epochs causes reconstruction loss $\mathcal{L}_{AE}$ to spike sharply. For lower-dimensional latent representations, reconstruction loss never recovers from this initial dip. However, with higher-dimensional representations, the loss begins to steadily decrease before approaching a steady state. In the case of $d = 4096$, loss values lower than those prior to introduction of the operator loss function can be observed.

\begin{figure}
  \includegraphics[width=\linewidth]{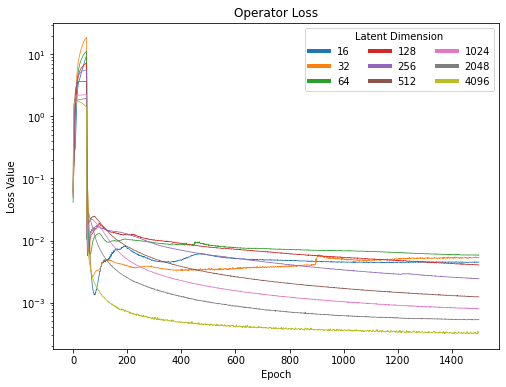}
  \caption{Operator loss as a function of latent dimension}
  \label{fig:operator_loss}
\end{figure}

Figure \ref{fig:operator_loss} shows results for operator loss. As opposed to reconstruction loss, we can observe operator loss rising significantly as the autoencoder network weights are updated during the first 50 training epochs. When operator loss is introduced, loss quickly drops before plateauing. Lower-dimensional latent representations show some variation in the loss surface. This is a result of the two loss functions both affecting the encoder weights. As the encoder learns latent representations with low reconstruction error, it also simultaneously find representations with a high degree of linearity.

Our results demonstrate a strong inverse correlation between latent dimensionality and loss values. High-dimensional latent representations result in models which outperform their low-dimensional analogs by several orders of magnitude. In all cases, minimum operator loss after training is at least a single order of magnitude lower than initial operator loss. However, introduction of the operator loss function significantly limits the performance of the autoencoder when latent dimensionality is low.

\section{Conclusion}

This work demonstrates the application of a series of invertible transformations to time-series volumetric data to allow an autoencoder to learn efficient latent representations. Graph canonization is shown to be an efficient alternative to graph similarity metrics like MPM during the process of volumetric graph reconstruction.  We conclude from our results that high-dimensional latent representations display strong linearity and show lowest values for both autoencoder and timestep operator loss metrics. Learned linear timestep operators show minimal loss when applied to vectors in the latent space.

Natural extensions to this work could involve assessing the effects of repeated applications of the linear timestep operator, removing linearity contraints on the timestep operator, testing with other types of materials, and testing with other types of physical processes in addition to mechanical deformation. The model can also be extended to high-dimensional data; while the architecture was developed with molecular data in mind, it should be possible to find a mapping to a representative latent space for any arbitrary set of time-series data.

\nocite{*}
\bibliographystyle{Jornadas}
\bibliography{biblio}

\end{document}